\documentclass{article}

\PassOptionsToPackage{numbers, compress}{natbib}
\usepackage[preprint]{neurips_2026}

\makeatletter
\renewcommand{\@noticestring}{}
\makeatother

\usepackage[utf8]{inputenc} 
\usepackage[T1]{fontenc}    
\usepackage{hyperref}       
\usepackage{url}            
\usepackage{booktabs}       
\usepackage{amsfonts}       
\usepackage{nicefrac}       
\usepackage{microtype}      
\usepackage{xcolor}         
\usepackage{graphicx}       

\usepackage{amsmath}        
\usepackage{algorithm}
\usepackage{algpseudocode}
\usepackage{pifont}         

\title{Dreamer-SAC: Off-Policy Learning in Latent World Models for Sample-Efficient Autonomous Driving}

\author{
  Jiazhuo Li \quad Linjiang Cao \quad Qi Liu \quad Xi Xiong\thanks{Corresponding author.} \\
  Tongji University \\
  \texttt{2534457@tongji.edu.cn}, \ \texttt{2431743@tongji.edu.cn}, \\
  \texttt{liu\_qi@tongji.edu.cn}, \ \texttt{xi\_xiong@tongji.edu.cn}
}

\begin{document}

\maketitle

\begin{abstract}
Sample-efficient reinforcement learning for autonomous driving is often limited by the trade-off between data efficiency and model bias. While world models reduce the reliance on costly environment interactions, policy optimization over learned dynamics remains sensitive to prediction errors. This paper proposes the Dreamer-SAC framework, which integrates a recurrent state-space world model with an off-policy soft actor-critic algorithm trained directly in latent space. The framework uses a combination of real interactions and short-horizon generated trajectories with n-step target estimation and multi-objective supervision. Evaluated in autonomous driving scenarios with objectives encompassing driving efficiency and safety, the proposed framework consistently outperforms representative reinforcement learning baselines, including DreamerV3, SAC, and PPO, while achieving improved performance with substantially fewer real environment interactions. Experiments reveal an inverted-U relationship between rollout horizon and policy performance, where short-horizon latent rollouts achieve the best trade-off between additional training signals and accumulated model bias. Furthermore, n-step target estimation demonstrates more effectiveness over one-step temporal-difference targets in exploiting predicted experience for value learning. 
\end{abstract}

\section{Introduction}\label{sec:intro}

Autonomous driving is a core enabling technology for next-generation intelligent transportation systems, with the potential to reduce traffic fatalities, alleviate congestion, and improve mobility accessibility \cite{fagnant2015preparing}. Autonomous vehicles are required to make continuous sequential decisions in complex and uncertain traffic environments involving dynamic interactions with surrounding road users. However, the large-scale deployment of reliable autonomous driving decision systems remains challenging. A primary bottleneck lies in the prohibitive cost of safety validation, as ensuring reliable performance under long-tail and safety-critical scenarios, such as collisions, unexpected cut-ins, and complex interactions, requires extensive driving experience. Real-world testing not only incurs substantial economic costs but also introduces potential risks to public road safety \cite{dona2022virtual}. Reinforcement learning (RL) provides a promising data-driven framework for sequential decision-making \cite{kiran2021deep}, enabling autonomous agents to learn sophisticated driving policies through environmental interaction rather than relying solely on manually designed decision rules. Despite its potential, the practical application of RL to autonomous driving remains severely constrained by its limited sample efficiency, as achieving reliable policies typically requires large amounts of interaction data \cite{zhao2024survey}.


Model-based reinforcement learning (MBRL) has recently attracted considerable attention as a promising solution to this problem. Instead of interacting with the environment for every policy update, MBRL learns a predictive world model that approximates environment dynamics and generates trajectories for policy optimization. By replacing expensive environment interactions with latent-space rollouts, world models substantially improve data efficiency while maintaining competitive control performance \cite{feng2025survey}. Building on this paradigm, this paper proposes a model-based RL framework tailored for autonomous driving. Our method integrates world models with off-policy RL, enabling the agent to learn policies through both real and rolled-out interactions. This maximizes data utilization while mitigating the impact of model bias.

Existing approaches to autonomous driving decision making have evolved through several paradigms. Early systems relied predominantly on rule-based methods, such as finite state machines, behavior trees, and handcrafted heuristics for scenario decomposition and maneuver selection \cite{gonzalez2016review,schwarting2018planning}. While interpretable and deterministic, these methods struggle to generalize to the combinatorial complexity of real-world traffic interactions and require extensive manual engineering to scale. Optimization-based techniques---notably model predictive control (MPC)---were subsequently introduced to explicitly exploit vehicle dynamics and constraints for trajectory planning and tracking \cite{falcone2007predictive}. However, the reliance on manually designed cost functions and simplified dynamic models limits their adaptability in highly interactive or uncertain environments. The emergence of deep learning enabled a paradigm shift toward data-driven policy learning. Imitation learning (IL), particularly behavior cloning, learns driving policies directly from expert demonstrations by mapping raw sensor inputs to control commands \cite{lemero2022survey}. Although effective for capturing human-like driving styles, IL fundamentally suffers from covariate shift and compounding errors when the learned policy deviates from the expert state distribution \cite{codevilla2018endtoend}. Reinforcement learning (RL) offers an alternative by enabling agents to discover policies through trial-and-error interaction, thereby optimizing long-term performance beyond expert demonstrations. Model-free RL algorithms, such as Proximal Policy Optimization (PPO) \cite{schulman2017proximal} and Soft Actor-Critic (SAC) \cite{haarnoja2018soft}, have achieved remarkable success in continuous control tasks. To improve sample efficiency, model-based methods learn predictive dynamics models that generate synthetic rollouts, reducing reliance on costly real interactions. Early work such as Model-Based Policy Optimization (MBPO) \cite{janner2019mbpo} demonstrated that short-horizon rollouts from an ensemble of probabilistic dynamics models can effectively augment real data for off-policy SAC training. However, MBPO operates directly in the original state space using explicit dynamics models, limiting its applicability to high-dimensional visual observations common in autonomous driving. To overcome this, latent world models learn compact representations of environment dynamics, enabling planning and policy optimization entirely in a compressed latent space \cite{hafner2019learning,hafner2019dream}. Among recent approaches, the Dreamer family has demonstrated remarkable success across continuous-control benchmarks by learning a Recurrent State-Space Model (RSSM) that captures latent dynamics \cite{hafner2025mastering}. The learned latent representation enables planning and policy learning entirely in latent space, avoiding costly pixel-level reconstruction during policy optimization.


Several works have adapted latent world models to autonomous driving: Gao et al.\ enhanced robustness through semantic masked world models \cite{gao2024enhance}, while Yang et al.\ explored representation alignment between raw observations and privileged features \cite{yang2025raw2drive}. However, these approaches primarily focus on improving the representation quality of the world model itself, rather than reconsidering the policy optimization paradigm within the latent space. Most of these methods employ on-policy actor-critic optimization strategies, limiting the agent to training within the ``imagined'' world and lacking the ability to learn directly from real-world data. Unlike real transitions, predicted trajectories inevitably suffer from accumulated prediction errors \cite{lambert2020challenges}, causing model bias that increases rapidly with rollout horizon, which will affect the policy's assessment of safety-critical driving scenarios (such as collisions).

To address these challenges, this paper proposes a Dreamer-SAC framework that combines latent world models with off-policy reinforcement learning through short-horizon rollouts and mixed replay optimization. Instead of relying exclusively on predicted trajectories, the proposed framework continuously updates the world model using real interaction data while generating short latent rollouts from posterior RSSM states. Real transitions are optimized using conventional one-step temporal-difference targets, whereas predicted trajectories adopt n-step targets to better exploit long-term reward information within the rollout horizon. Both data sources are jointly used for critic and actor optimization, allowing SAC to benefit from model-generated experience while limiting the adverse effects of long-horizon model bias.

The main contributions of this work are summarized as follows:

\begin{enumerate}
\item A model-based off-policy reinforcement learning framework is proposed by integrating Soft Actor-Critic into the Recurrent State-Space Model, enabling autonomous driving policies to efficiently exploit model-generated rollouts.

\item A hybrid real and model-generated experience learning strategy is developed to improve policy reliability by combining real transitions and latent rollouts with different value estimation schemes, thereby enhancing data utilization while mitigating the impact of model prediction errors.

\item Comprehensive experiments demonstrate the superiority of the proposed framework over representative reinforcement learning baselines and reveal an inverted-U relationship between rollout horizon and performance, indicating that short-horizon rollouts provide the optimal balance between learning benefits and model bias.
\end{enumerate}

\section{Preliminaries}\label{sec:preliminaries}

This section introduces the problem we study, as well as background knowledge regarding world models and Soft Actor-Critic.

\subsection{Problem Definition}

In real-world driving, an ego vehicle cannot access the complete and exact state of the environment, such as the precise intentions of other drivers or occluded road regions. Instead, it must rely solely on its onboard sensors, which provide partial and noisy observations. We model the autonomous driving task as a Partially Observable Markov Decision Process (POMDP) \cite{kaelbling1998planning}. The POMDP is defined by the tuple $(\mathcal{S}, \mathcal{A}, \mathcal{O}, \mathcal{T}, \Omega, \mathcal{R}, \gamma)$, where  $\mathcal{S}$ represents the true but unobservable environment state. $\mathcal{A}$ is the action space consisting of continuous steering and throttle/brake commands $\mathbf{a}_t = (a_{\text{steer}}, a_{\text{throttle}}) \in [-1, 1]^2$. $\mathcal{O}$ is the observation space. At each time step $t$, the ego vehicle receives an observation $\mathbf{o}_t$ composed of two modalities:
\begin{equation}
\mathbf{o}_t = \{I_t, \mathbf{v}_t\},
\end{equation}
where $I_t \in \mathbb{R}^{H \times W \times 3}$ is an image from a front-facing camera, and $\mathbf{v}_t \in \mathbb{R}^{125}$ is a vector of vehicle observation, including LiDAR and own-state information. Both of them can be directly obtained through raw sensors or physical calculations. $\mathcal{T}(s' \mid s, a): \mathcal{S} \times \mathcal{A} \times \mathcal{S} \rightarrow [0, 1]$ defines the conditional transition probability distribution over next states. $ \Omega (o \mid s, a): \mathcal{S} \times \mathcal{A} \times \mathcal{O} \rightarrow [0, 1]$ defines the probability of observing given state and action. $\mathcal{R}: \mathcal{S} \times \mathcal{A} \rightarrow \mathbb{R}$ is the reward function balancing safety and efficiency, while $\gamma \in [0, 1]$ is the discount factor. The agent's objective is to learn a policy $\pi(a_t | o_{\leq t})$ that maximizes the expected cumulative discounted reward:
\begin{equation}
\mathbb{E}_{\pi} \left[ \sum_{t=0}^{\infty} \gamma^t r_t \right].
\end{equation}

\subsection{Latent World Models}

Model-based reinforcement learning improves sample efficiency by learning a predictive model of environment dynamics and generating trajectories for policy optimization. Instead of directly predicting future observations in the high-dimensional image space, latent world models first compress observations into a compact latent representation $\hat{s}_{t} \in \mathcal{S}_{\text{latent}}$ and perform transition prediction entirely in the latent space. This substantially reduces computational cost while preserving information relevant for control.

The world model consists of the following components:

\begin{align}
\text{Representation model:}&\quad \hat{s}_t \sim p(\hat{s}_t \mid \hat{s}_{t-1}, a_{t-1}, o_t) \\
\text{Transition model:}&\quad \hat{s}_{t+1} \sim p(\hat{s}_{t+1} \mid \hat{s}_t, a_t) \\
\text{Observation model:}&\quad \hat{o}_t \sim p(\hat{o}_t \mid \hat{s}_t) \\
\text{Reward model:}&\quad \hat{r}_t \sim p(\hat{r}_t \mid \hat{s}_t, a_t) \\
\text{Continue model:}&\quad \hat{c}_t \sim p(\hat{c}_t \mid \hat{s}_t, a_t)
\end{align}

The representation model maps high-dimensional observations into a compact latent state, compressing raw sensor inputs while preserving task-relevant information. The transition model predicts future latent states given the current state and action, forming the core of the learned dynamics. The observation model reconstructs observations from the latent state, providing a self-supervised reconstruction signal. The reward model predicts immediate rewards for value estimation. In our tasks, a separate continue model predicts the termination signal, helping the agent identify episode boundaries. Through the learned latent dynamics model, the agent can generate latent rollouts by predicting future states, rewards, and termination signals in the latent space. These model-generated trajectories provide additional training experiences for policy optimization and reduce the reliance on direct environment interactions.

\subsection{Soft Actor-Critic}

Soft Actor-Critic (SAC) \cite{haarnoja2018soft} is an off-policy actor-critic algorithm that optimizes both expected return and policy entropy, encouraging efficient exploration while maintaining stable learning. SAC solves the maximum entropy objective:
\begin{equation}
J(\pi) = \sum_{t=0}^{\infty} \mathbb{E}_{(s_t, a_t) \sim \rho_\pi} \Bigl[ \gamma^t r(s_t, a_t) + \alpha \mathcal{H}\bigl(\pi(\cdot \mid s_t)\bigr) \Bigr],
\end{equation}
where $\rho_\pi$ denotes the state-action marginal induced by policy $\pi$, $\gamma \in [0,1)$ is the discount factor, and $\alpha$ controls the trade-off between reward maximization and the entropy term $\mathcal{H}$, which encourages exploration. The critic is trained using the soft Bellman target:
\begin{equation}
y_t = r_t + \gamma \,\mathbb{E}_{a_{t+1} \sim \pi_\phi}\Bigl[ \min_{i=1,2} Q_{\bar{\theta}_i}(s_{t+1}, a_{t+1}) - \alpha \log \pi_\phi(a_{t+1} \mid s_{t+1}) \Bigr],
\end{equation}
where two independent $Q$-networks are employed to alleviate overestimation bias. The policy is optimized by minimizing the actor loss:
\begin{equation}
\mathcal{L}_\pi(\phi) = \mathbb{E}_{s_t \sim \mathcal{D}}\; \mathbb{E}_{a_t \sim \pi_\phi} \Bigl[ \alpha \log \pi_\phi(a_t \mid s_t) - \min_{i=1,2} Q_{\theta_i}(s_t, a_t) \Bigr],
\end{equation}
where $\mathcal{D}$ is the replay buffer and the temperature parameter $\alpha$ is automatically adjusted to maintain a target policy entropy. Compared with on-policy policy-gradient methods, SAC achieves substantially higher sample efficiency through experience replay and stable value estimation, making it well suited for latent model-based reinforcement learning.

\section{Method}\label{sec:method}

\begin{figure}[htb]
\centering
\includegraphics[width=\linewidth]{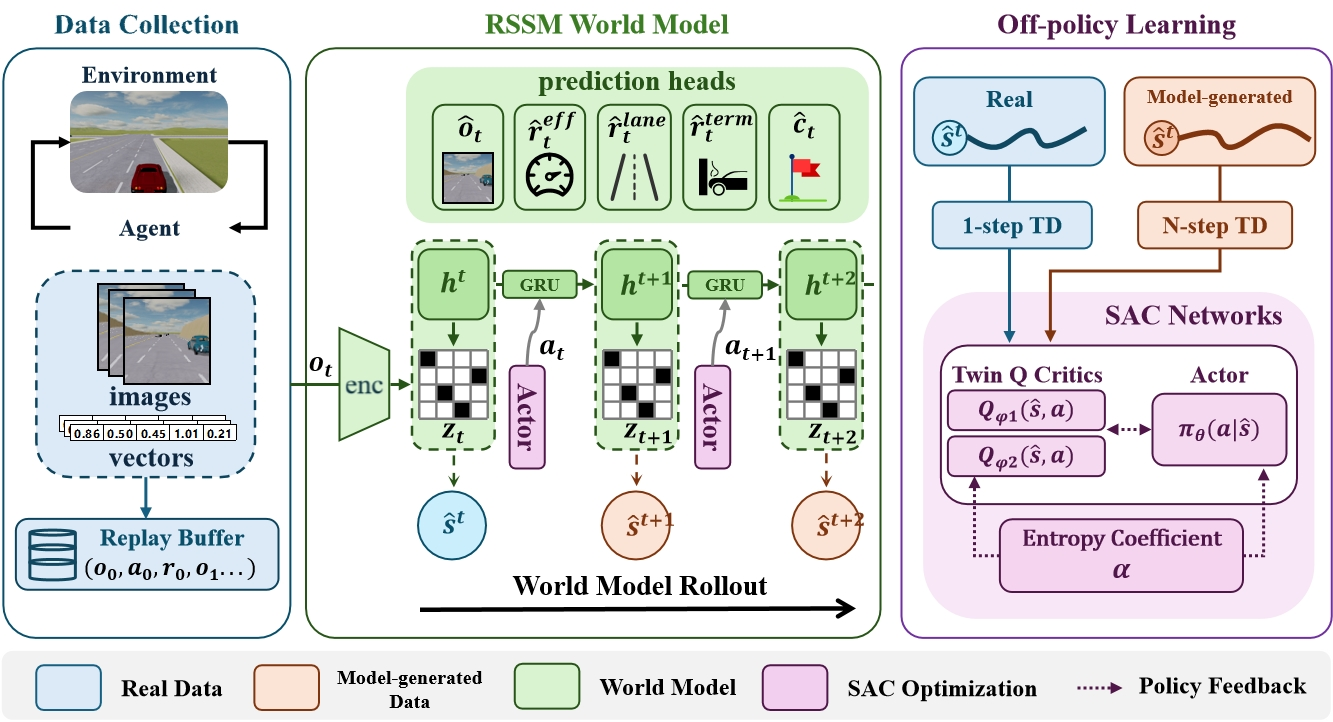}
\caption{Overview of the proposed Dreamer-SAC framework. The RSSM encodes multimodal observations into a compact latent state; latent rollouts are generated from posterior states and jointly used with real transitions to optimize the SAC policy.}\label{fig:method_framework}
\end{figure}

This section introduces the Dreamer-SAC framework. As shown in Figure~\ref{fig:method_framework}, real driving experiences collected from the environment are stored in a replay buffer and used to train the RSSM world model, which learns compact latent representations and predicts observations, multi-objective rewards, and termination signals. Starting from real latent states, the current policy performs short-horizon rollouts in the learned latent dynamics. The generated trajectories are combined with real experiences to train the SAC agent, where real data adopts one-step temporal-difference targets and model-generated data uses multi-step returns. By integrating real and predicted experiences within an off-policy learning framework, the proposed method improves data efficiency while reducing the model bias associated with purely model-generated policy optimization.

\subsection{Multi-modal Encoding}
The observation at each timestep consists of two modalities: a front-facing camera image $I_t \in \mathbb{R}^{H \times W \times 3}$ providing rich visual context of the road environment, and a vector $\mathbf{v}_t \in \mathbb{R}^{125}$ comprising LiDAR range measurements and ego-vehicle state information. The LiDAR component captures 120 range readings over a 50-meter sensing radius, while the ego-state component encodes the vehicle's speed, steering angle, previous actions and yaw rate.

The image modality is processed by a convolutional neural network (CNN): \( f_{\text{img}} = \text{CNN}(I_t; \theta_{\text{img}}) \), progressively downsampling spatial resolution while expanding channel depth. The resulting feature maps are flattened into a compact image embedding. The vector modality is processed by a multi-layer perceptron (MLP): \( f_{\text{phys}}= \text{MLP}(v_t; \theta_{\text{vec}}) \). Both embeddings are then concatenated to form the encoded observation:
\begin{equation}
e_t = \text{Concat}(f_{\text{img}}, f_{\text{phys}}) \in \mathbb{R}^{d_{\text{img}} + d_{\text{phys}}}.
\end{equation}

\subsection{Latent Dynamics Modeling}

The encoded observations embedding is then fed into a Recurrent State-Space Model (RSSM) \cite{hafner2025mastering}. At each time step, the model maintains a deterministic hidden state $h_t$ that summarizes past information, and a stochastic state $z_t$ that captures uncertainty. The transition is defined as:
\begin{equation}
\begin{aligned}
&h_t = f_\theta(h_{t-1}, z_{t-1}, a_{t-1}), \\
&\text{Prior: } \hat{z}_t \sim p_\theta(\hat{z}_t \mid h_t), \\
&\text{Posterior: } z_t \sim q_\theta(z_t \mid h_t, e_t),
\end{aligned}
\end{equation}
where $f_\theta$ is a recurrent neural network. $z_t$ is sampled from the posterior $q(z_t \mid h_t, e_t)$ when real observations are available, and from the prior $p(z_t \mid h_t)$ during latent rollouts where no observation exists. $(h_{t}, z_{t})$ together form the latent state.

The basic world model loss consists of prediction loss and KL regularization loss:
\begin{equation}
\label{eq:wm_basic}
\mathcal{L}_{\text{basic}} = \mathbb{E}_{q_\theta}\left[\sum_{t=1}^{T} \left(\mathcal{L}_{\text{pred},t} + \mathcal{L}_{\text{KL},t}\right)\right],
\end{equation}
where $\mathcal{L}_{\text{pred},t}$ includes the negative log-likelihood of observation reconstruction $-\ln p_\theta(o_t \mid h_t, z_t)$, reward prediction $-\ln p_\theta(r_t \mid h_t, z_t, a_t)$, and termination signal prediction $-\ln p_\theta(c_t \mid h_t, z_t, a_t)$. The $\mathcal{L}_{\text{KL}, t} = D_{\text{KL}}\big(q_\theta(z_t \mid h_t, e_t) \parallel p_\theta(\hat{z}_t \mid h_t)\big)$ is the KL divergence between the posterior and prior distribution, constraining the encoder to avoid extracting irrelevant information from observations and enhancing the consistency and predictability of the latent state.

\subsection{Multi-objective Reward}

Autonomous driving is inherently a multi-objective decision-making problem, where safe and efficient driving behaviors must be optimized simultaneously. The reward function consists of the following three parts:

The driving efficiency reward $R_{\text{eff}}$ encouraging forward progress and maintaining appropriate speed:
\begin{equation}
R_{\text{eff}} = \beta_d \times (x_t - x_{t-1}) \times \mathbf{1}_{\text{pos}}+\beta_s \times (v / v_{\max}) \times \mathbf{1}_{\text{pos}},
\end{equation}
where $x_t$ is the longitudinal projection of the vehicle onto the lane centerline, $v$ is the current speed, $v_{\max}$ is the expected maximum speed, and $\mathbf{1}_{\text{pos}}$ indicates correct driving direction.

The lane center offset penalty $R_{\text{lane}}$ penalizes deviation from the lane center:
\begin{equation}
R_{\text{lane}} = -\beta_c \times (|y_t|) / w_{\text{lane}},
\end{equation}
where $y_t$ denotes lateral offset and $w_{\text{lane}}$ is lane width.

The termination reward/punishment $R_{\text{term}}$ provides a sparse terminal reward for completing the route or penalty for violations:
\begin{equation}
R_{\text{term}} =
\begin{cases}
r_{\text{success}} & \text{if episode completes successfully} \\
-p_{\text{crash}} & \text{if collision occurs} \\
-p_{\text{out}} & \text{if vehicle drives out of road} \\
\end{cases}.
\end{equation}

The total reward for each timestep is:
\begin{equation}
R = R_{\text{eff}} + R_{\text{lane}} + R_{\text{term}}.
\end{equation}

Instead of learning a single reward predictor, we employ three independent prediction heads. The efficiency reward $R_{\text{eff}}$ and lane-center reward $R_{\text{lane}}$ are continuous and are optimized using mean squared error (MSE) loss. In contrast, the terminal reward $R_{\text{term}}$ is sparse and discrete. A symlog-discretized categorical distribution is used to model $R_{\text{term}}$, which better captures the uncertainty of terminal events by representing predictions as a probability distribution over discrete bins.

The overall reward prediction loss is defined as:
\begin{equation}
\label{eq:reward_loss}
\mathcal{L}_{\text{reward}} = \lambda_1 \mathcal{L}_{\text{MSE}}(R_{\text{eff}}, \hat{R}_{\text{eff}}) + \lambda_2 \mathcal{L}_{\text{MSE}}(R_{\text{lane}}, \hat{R}_{\text{lane}}) + \lambda_3 \mathcal{L}_{\text{CE}}(R_{\text{term}}, \hat{R}_{\text{term}}),
\end{equation}
where $\mathcal{L}_{\text{CE}}$ denotes the cross-entropy loss for the symlog-discretized categorical prediction, and $\lambda_i$ are weighting coefficients balancing the contribution of each reward component. Therefore, the reward prediction term in the world model loss (Equation~\ref{eq:wm_basic}) is replaced by $\mathcal{L}_{\text{reward}}$ (Equation~\ref{eq:reward_loss}).

\subsection{Off-policy Learning}

In Dreamer-style methods, the actor-critic is trained on-policy:
each update requires fresh trajectories generated by the current
policy, which is why these methods rely exclusively on
model-generated rollouts and cannot reuse previously collected
real transitions, as those were produced by earlier policies.
In contrast, off-policy reinforcement learning imposes no such
restriction on the data-generating policy. This allows the agent
to simultaneously exploit stored real transitions and freshly
generated latent rollouts, enabling more efficient reuse of
expensive real-world samples while still benefiting from
model-generated experiences.

\subsubsection{Hybrid Real and Model-Generated Replay}

The proposed hybrid replay mechanism operates as follows. For each training iteration, a batch of trajectory segments is first sampled from the real replay buffer, $\mathcal{D}_r = \{(o_t, a_t, r_t, \cdots)\}$. These real trajectories serve two purposes. First, they are used to optimize the RSSM world model through representation learning. Second, the posterior latent states inferred from each real transition are used as initialization points for latent rollouts. Starting from every posterior state, the world model rolls forward for a fixed rollout horizon $H$ using the current SAC policy, generating transitions $\mathcal{D}_i$. The SAC agent is then updated using the union of real and model-generated experiences, $\mathcal{D} = \mathcal{D}_r \cup \mathcal{D}_i$.

Since rollout is performed from every sampled real transition, the number of predicted transitions is directly determined by the rollout horizon $H$. Consequently, the ratio between real and predicted samples automatically becomes $1:H$. Unlike previous methods that manually tune the proportion of predicted data, the proposed strategy introduces no additional hyperparameter. Increasing the rollout horizon naturally increases the amount of model-generated supervision, reflecting greater reliance on the learned world model.

Moreover, predicted trajectories are discarded immediately after policy optimization. At the next iteration, new predicted samples are regenerated using the latest RSSM and SAC policy. This online regeneration avoids repeatedly training on outdated predictions and reduces the influence of model errors accumulated during previous optimization stages.

\subsubsection{Multi-step Bootstrapping}

Real transitions correspond to actual environment interactions and therefore employ the standard one-step SAC target,
\begin{equation}
\label{eq:td1_real}
y_t^{\text{real}} = r_t + \gamma \, c_t \Bigl( \min_{i=1,2} Q_{\bar{\theta}_i}(s_{t+1}, a_{t+1}) - \alpha \log \pi_\phi(a_{t+1} \mid s_{t+1}) \Bigr), \quad a_{t+1} \sim \pi_\phi(\cdot \mid s_{t+1}),
\end{equation}
where $c_t \in \{0, 1\}$ is the continuation flag, which zeroes out the bootstrap term after collision, out-of-road, or successful arrival. In contrast, predicted trajectories are generated entirely inside the learned world model and contain future rewards that are immediately available after rollout. Therefore, instead of discarding these additional rewards, we exploit the complete rollout horizon by constructing an $n$-step target,
\begin{equation}
\label{eq:td_n_pred}
y_t^{\text{pred}} = \sum_{k=0}^{N-1} \gamma^k \Bigl( \prod_{j=0}^{k-1} c_{t+j} \Bigr) r_{t+k} + \gamma^N \Bigl( \prod_{j=0}^{N-1} c_{t+j} \Bigr) \Bigl( \min_{i=1,2} Q_{\bar{\theta}_i}(s_{t+N}, a_{t+N}) - \alpha \log \pi_\phi(a_{t+N} \mid s_{t+N}) \Bigr),
\end{equation}
where $a_{t+N} \sim \pi_\phi(\cdot \mid s_{t+N})$, $N$ equals the remaining length of the rollout, and $c_{t+j}$ is from the RSSM continue predictor. Compared with one-step TD learning, the proposed target propagates long-horizon reward information to the critic while reducing bootstrap variance inside predicted trajectories. Algorithm~\ref{alg:training_v2} summarizes the complete training procedure.

\begin{algorithm}[htb]
\caption{Dreamer-SAC: Hybrid World Model with Off-policy Policy Optimization}
\label{alg:training_v2}
\begin{algorithmic}[1]

\State \textbf{Input:} World model $M$, actor $\pi$, critics $Q_1, Q_2$, real replay buffer $\mathcal{D}$, rollout horizon $H$

\State Initialize $M, \pi, Q_1, Q_2$

\While{training}

    \State Collect one transition from the environment and store it in $\mathcal{D}$

    \State Sample batch $\{(o_t, a_t, r_t)\}_{t=1}^{T}$ from $\mathcal{D}$

    \State Update RSSM by minimizing $\mathcal{L}_{\text{WM}} = \mathcal{L}_{\text{KL}} + \mathcal{L}_{\text{obs}} + \mathcal{L}_{\text{reward}} + \mathcal{L}_{\text{cont}}$

    \State Initialize predicted buffer $\hat{\mathcal{D}} \leftarrow \emptyset$

    \ForAll{posterior latent state $s_t$}
        \State $s \leftarrow \text{sg}(s_t)$  \Comment{Stop gradient}
        \For{$k = 1$ \textbf{to} $H$}
            \State Sample action $a_k \sim \pi(s)$
            \State Predict $(s', r, c)$ using RSSM
            \State Store transition $(s, a, r, c, s')$ into $\hat{\mathcal{D}}$
            \State $s \leftarrow s'$
        \EndFor
    \EndFor

    \State Update critics using real samples with 1-step TD targets and predicted samples with $n$-step targets

    \State Update actor by SAC objective

    \State Update entropy coefficient $\alpha$

    \State Soft-update target critics

    \State Discard $\hat{\mathcal{D}}$

\EndWhile

\end{algorithmic}
\end{algorithm}

\section{Experiments}\label{sec:experiments}

This section provides a detailed introduction to our experimental setup, model configuration, comparative experiment, ablation study and analysis of related results.

\subsection{Experimental Setup}

We evaluate the proposed framework on the MetaDrive simulator \cite{li2022metadrive}, a high-fidelity autonomous driving platform that provides diverse road geometries and dynamic traffic interactions. The experiments are conducted on the BIG\_BLOCK\_SEQUENCE-CC scenario, which consists of a multi-lane highway segment with consecutive curved sections. The vehicle itself needs to continuously adjust its steering as the road curvature changes, while maintaining safe interaction with surrounding vehicles.

Key environment parameters include traffic density of 0.1 (Approximately 30 vehicles per kilometer), 10 distinct training scenarios and 50 distinct testing scenarios. Each episode terminates immediately upon collision or out-of-road. Each episode runs for up to 300 decision steps, with each decision step corresponding to 20 simulation frames. The action space is $[-1, 1]^2$ for continuous steering and throttle/brake control. Observations consist of an RGB camera image ($84 \times 84 \times 3$) and a 125-dimensional vector (120 LiDAR rays at 50\,m range plus 5 ego-vehicle state variables). Collision and out-of-road penalty $p_{crash}, p_{out} = 40.0$, successful arrival bonus $r_{success} = 10.0$. Reward coefficients $\beta_d = 1.0$, $\beta_s = 0.1$, $\beta_c = 1.0$.

\begin{figure}[htb]
\centering
\includegraphics[width=0.7\linewidth]{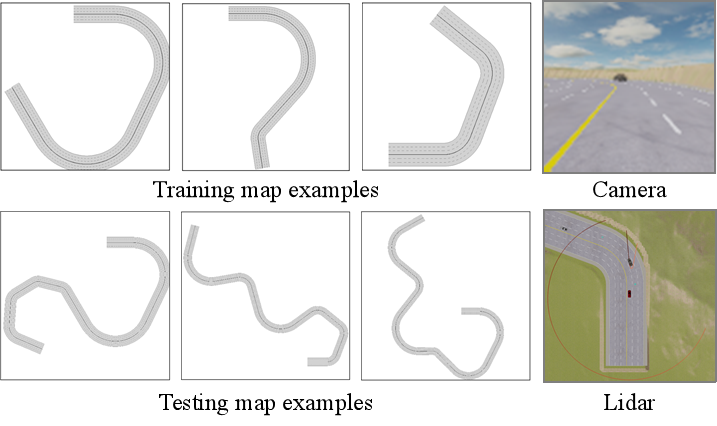}
\caption{MetaDrive driving scenario with multimodal observations.}
\label{fig:environment}
\end{figure}

\subsubsection{Model Configurations}

The world model is implemented based on the recurrent state-space model (RSSM) architecture. The image encoder consists of four convolutional layers with channel sizes of 32, 64, 128, and 256, respectively, using kernel size 4, stride 2, layer normalization, and SiLU activations. The vector observation encoder employs a two-layer MLP with 256 hidden units, layer normalization, and SiLU activations. The RSSM contains a deterministic recurrent state of 256 dimensions and a discrete stochastic state with 32 categorical variables and 32 classes per variable. The resulting latent feature dimension is 1280. The decoder, reward prediction heads, and continuation prediction head are optimized jointly with the RSSM to learn predictive latent representations.

The SAC policy module operates directly on RSSM latent features. The actor and twin critic networks consist of two fully connected layers with 256 hidden units, layer normalization, and SiLU activations. The critic networks use independent parameters and are updated with soft target networks using a Polyak coefficient of 0.005. The entropy coefficient is automatically adjusted during training with a target entropy of $-2$.

All networks are optimized using the Adam optimizer. The world model is trained with sequences of batch size 16 and sequence length 32, using a learning rate of $1\times10^{-4}$. The SAC actor and critic are optimized with a learning rate of $3\times10^{-4}$. The rollout horizon is set to $H=5$ steps, where predicted trajectories are generated from posterior RSSM states and combined with real transitions for off-policy learning. The discount factor is $\gamma=0.99$. Training is performed for 40,000 environment interaction steps. The overall model contains approximately 13 million trainable parameters, including the world model, SAC policy, and value networks.

\subsection{Comparative Experiments}

To evaluate the effectiveness of the proposed Dreamer-SAC framework, we compare it with three representative reinforcement learning baselines. All methods are evaluated under identical driving environments, including the same MetaDrive configuration, sensor inputs, training scenarios, and network capacity. For fair comparison, all actor-critic networks adopt the same architecture with two hidden layers of 256 units.

$\bullet$ \textbf{DreamerV3:} DreamerV3 \cite{hafner2025mastering} represents latent world-model-based on-policy learning, which learns a RSSM from interaction data and trains an actor-critic agent entirely on predicted latent rollouts generated by the learned model. The actor is updated via REINFORCE with an advantage estimate computed from predicted returns, while the critic predicts state values using a symlog-discretized distribution.

$\bullet$ \textbf{SAC:} Soft Actor-Critic is an off-policy maximum-entropy RL algorithm that jointly optimizes expected return and policy entropy. It employs twin Q-networks to mitigate overestimation bias, a replay buffer for sample reuse, and automatic entropy temperature tuning.

$\bullet$ \textbf{PPO:} Proximal Policy Optimization is an on-policy policy-gradient method that constrains policy updates via a clipped surrogate objective, balancing sample efficiency with training stability. PPO collects trajectories using the current policy, computes advantage estimates via generalized advantage estimation, and performs multiple epochs of minibatch updates on each collected batch before discarding it.

\begin{figure}[htbp]
\centering
\includegraphics[width=0.7\linewidth]{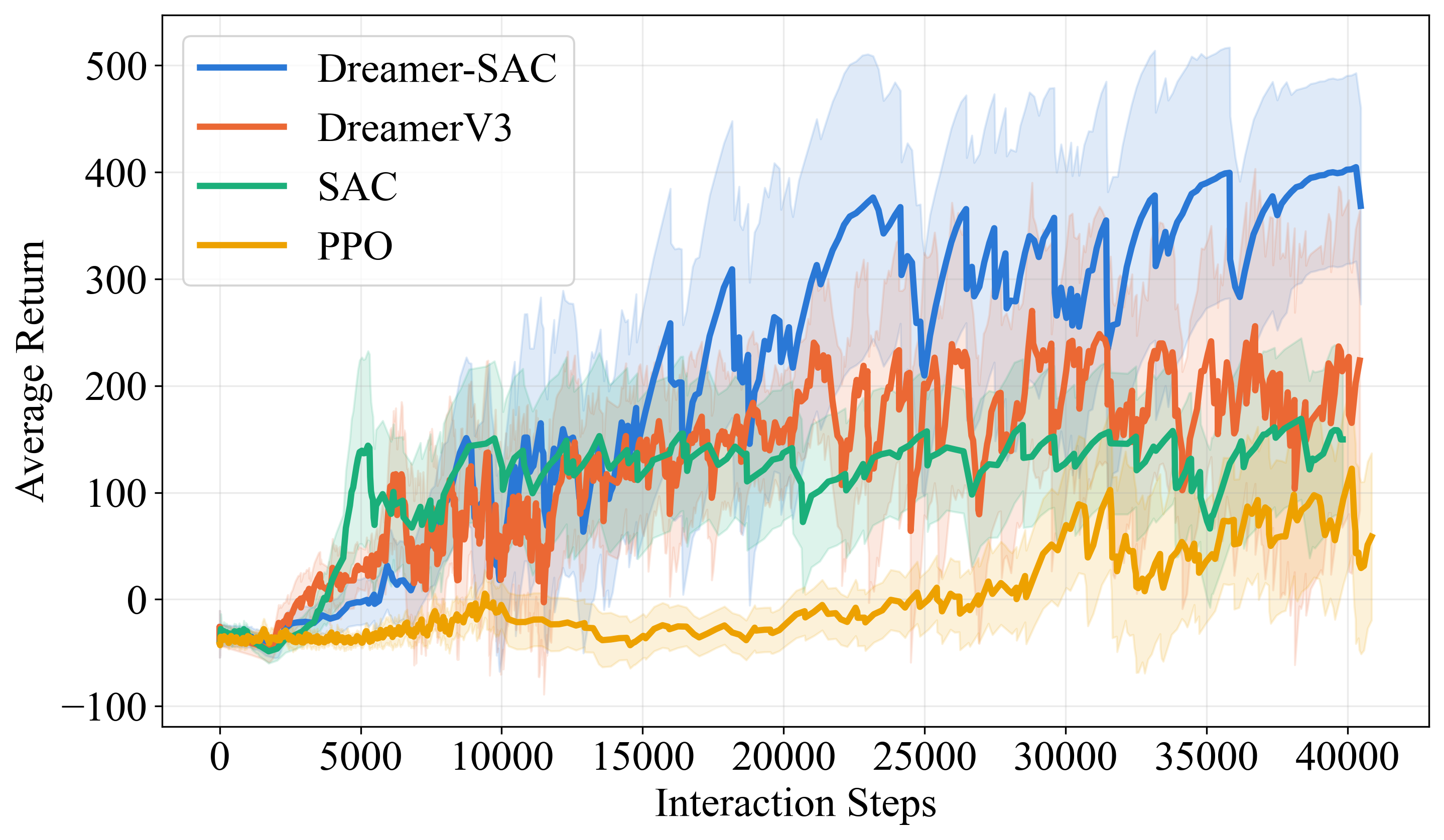}
\caption{Training return comparison of Dreamer-SAC against DreamerV3, SAC, and PPO baselines.}\label{fig:comparison}
\end{figure}

Figure~\ref{fig:comparison} shows the comparison curves of training return convergence. The proposed Dreamer-SAC achieves the highest final performance, reaching an average return of 371.4. In comparison, DreamerV3 achieves an average return of 189.2 and exhibits relatively high variance. The SAC curve is relatively stable, converging at 134.5. PPO employs online learning, resulting in low data efficiency and a slow-rising curve, reaching a score of 65.5. These results validate the effectiveness of integrating latent world models with off-policy policy optimization for autonomous driving tasks.

\subsubsection{Generalization Evaluation}

To further evaluate the generalization capability and long-horizon driving performance, we conduct additional experiments on a substantially longer road network, whose length is approximately three times that of the training environment. All methods are evaluated under 50 independent trials with unseen random seeds, and all indicators are taken as the average of these 50 experiments.

We measure five driving metrics, including collision frequency, out-of-road frequency, average speed, average travel distance, and maximum travel distance. Collision frequency and out-of-road frequency are normalized by the total driven distance:
\begin{equation}
\text{Collision Frequency} = \frac{N_{\text{collision}}}{D_{\text{total}}}, \qquad
\text{Out-of-road Frequency} = \frac{N_{\text{out}}}{D_{\text{total}}},
\end{equation}
where $N_{\text{collision}}$ and $N_{\text{out}}$ denote the number of collision and out-of-road events, respectively, and $D_{\text{total}}$ represents the total traveled distance in kilometers.

The results are summarized in Table~\ref{tab:generalization}. Dreamer-SAC achieves the lowest collision frequency (1.56/km) and out-of-road frequency (1.37/km), while also obtaining the longest average travel distance (320.8\,m) and maximum travel distance (951.6\,m). Compared with DreamerV3, SAC, and PPO, Dreamer-SAC demonstrates substantially improved long-horizon driving reliability, indicating that the integration of latent prediction with off-policy learning leads to more robust decision-making beyond the training scenarios.

\begin{table}[htbp]
\caption{Generalization performance on extended road networks.}\label{tab:generalization}
\centering
\begin{tabular}{lccccc}
\toprule
\textbf{Method} & \textbf{Collision} & \textbf{Out-of-road} & \textbf{Avg Speed} & \textbf{Avg Dist} & \textbf{Max Dist} \\
                & \textbf{(/km)}     & \textbf{(/km)}       & \textbf{(km/h)}    & \textbf{(m)}      & \textbf{(m)} \\
\midrule
Dreamer-SAC     & \textbf{1.56}   & \textbf{1.37}   & 25.5  & \textbf{320.8} & \textbf{951.6} \\
DreamerV3       & 2.88   & 1.76   & \textbf{39.7}  & 215.5 & 515.2 \\
SAC             & 1.68   & 8.85   & 28.0  & 95.0  & 348.0 \\
PPO             & 3.72   & 4.03   & 25.1  & 129.1 & 369.6 \\
\bottomrule
\end{tabular}
\end{table}


An interesting observation is that DreamerV3 achieves the highest average speed (39.7 km/h) but suffers from higher collision frequency and shorter travel distance. This indicates that training exclusively with predicted experiences may amplify model bias, where safety-critical transitions are inaccurately represented in the learned dynamics. Consequently, the policy may exploit inaccuracies in the predicted environment and adopt overly aggressive behaviors, such as maintaining higher speeds and performing frequent overtaking maneuvers, which improves short-term reward accumulation but compromises long-term driving reliability.

In contrast, Dreamer-SAC employs a dual-network architecture and learns a more conservative and reliable driving strategy by continuously incorporating real experiences during policy optimization. When encountering slower vehicles, the agent tends to follow the preceding vehicle and reduce speed when necessary rather than aggressively performing risky maneuvers. Although the average speed is lower than DreamerV3, this behavior significantly improves survival distance and reduces safety violations, which is particularly important for autonomous driving applications.

\subsection{Effect of Rollout Horizon}
\begin{figure}[H]
\centering
\includegraphics[width=0.50\linewidth]{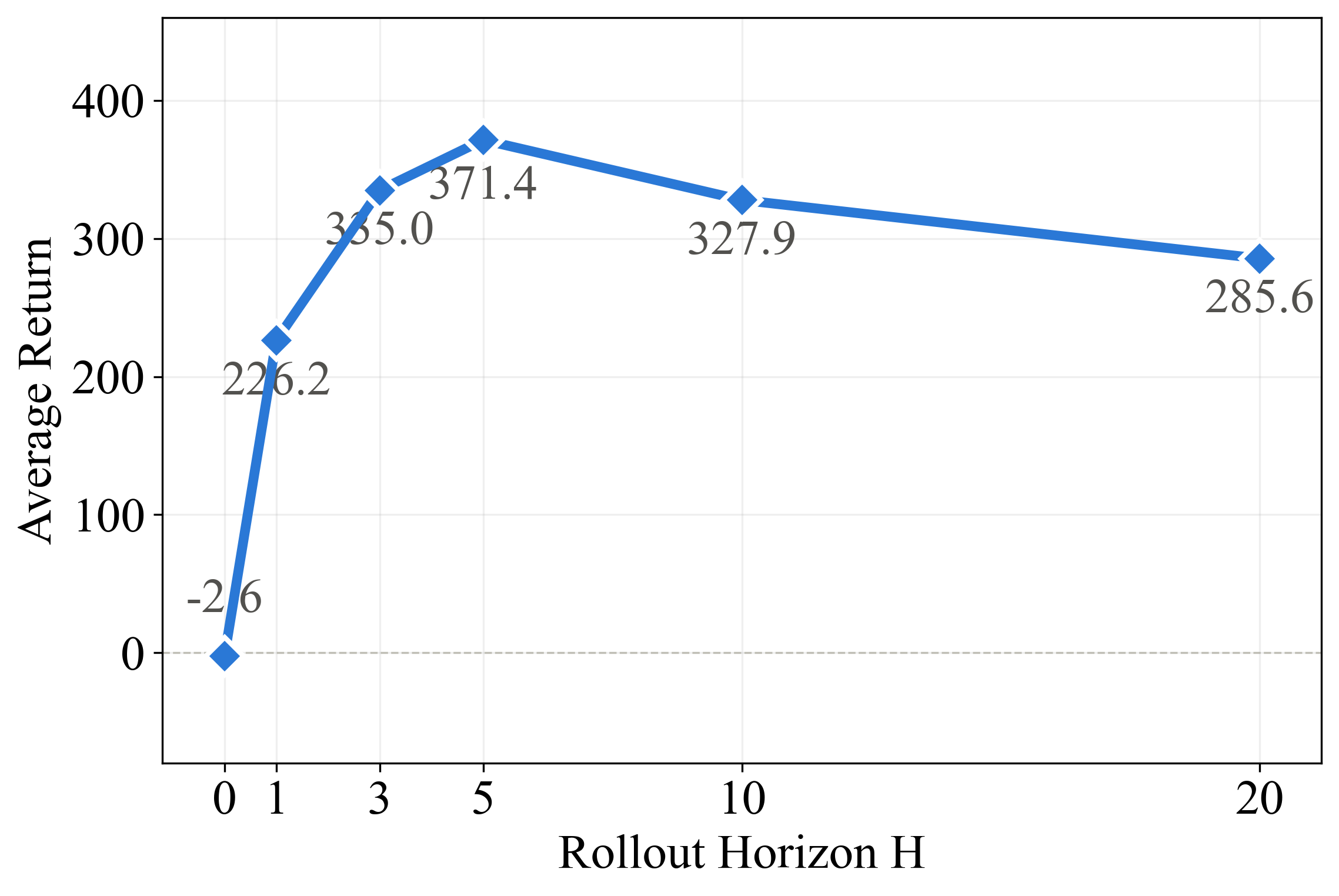}
\caption{Training return as a function of the rollout horizon $H$, showing an inverted-U trend with optimal performance at $H=5$.}\label{fig:horizon_ablation}
\end{figure}

The rollout horizon $H$ determines not only the rollout length of the learned world model but also the proportion of predicted samples used for policy optimization. We evaluate its influence by varying $H$ from 0 to 20 while keeping all other training configurations unchanged. As shown in Figure~\ref{fig:horizon_ablation}, the performance exhibits a clear inverted-U trend. Without rollouts ($H=0$), the agent fails to learn meaningful driving behavior and achieves an average return of only $-2.6$. Increasing the horizon significantly improves performance, reaching the best result at $H=5$ with an average return of 371.4. However, further increasing the horizon leads to gradual degradation, with returns decreasing to 285.6 at $H=20$.

The poor performance without rollouts indicates that directly optimizing SAC in the learned latent space using only real samples is insufficient during early training. Since the RSSM representation and policy are optimized separately, the initial critic receives limited and unstable decision-related information. As a result, the policy tends to converge to a conservative failure mode, where the vehicle remains stationary to avoid collision penalties. Introducing short-horizon rollouts provides additional on-policy samples generated by the current policy and world model, allowing the critic to continuously evaluate newly explored behaviors and avoiding the negative feedback loop caused by insufficient exploration.

\begin{figure}[H]
\centering
\includegraphics[width=0.95\linewidth]{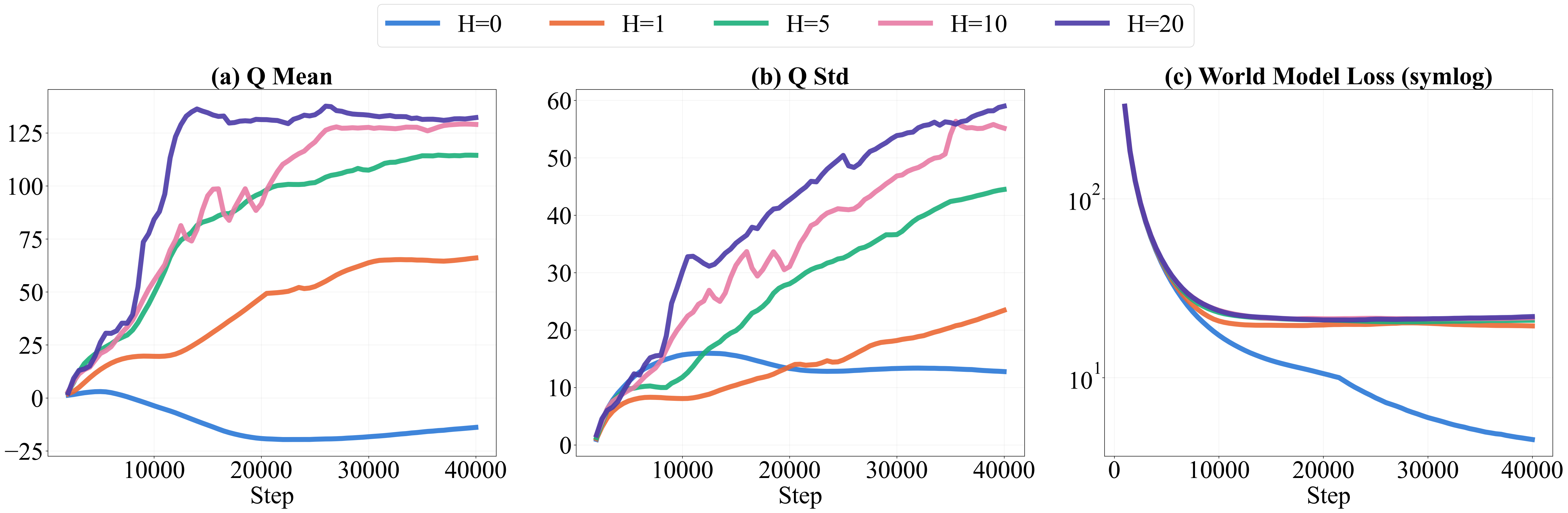}
\caption{Q-value mean, Q-value standard deviation, and world model loss across different rollout horizons.}\label{fig:metrics_comparison}
\end{figure}

The additional training statistics, illustrated in Figure~\ref{fig:metrics_comparison}, provide further insights into this phenomenon.  The average predicted Q-value increases with larger horizons, and the Q-value variance also becomes larger, suggesting that the critic obtains stronger action discrimination but also becomes increasingly optimistic. In particular, although $H=20$ achieves the highest Q-value estimates, its final return is lower than that of $H=5$, revealing a mismatch between value estimation and actual driving performance.

The world model loss also increases with longer rollout horizons. This does not indicate worse world model learning; instead, it reflects that longer rollouts expose the model to more diverse states generated by imperfect policies. In contrast, the extremely low loss at $H=0$ is mainly caused by the degenerated stationary policy, where the model only needs to reconstruct limited static observations. These results demonstrate that the rollout horizon introduces a fundamental trade-off between data augmentation and model bias: short rollouts provide reliable synthetic experiences and improve exploration, whereas excessive rollouts amplify prediction errors and lead to overestimated value functions. Therefore, $H=5$ provides the optimal balance between additional training signals and model reliability in our autonomous driving setting.

\subsection{Ablation study}

We conduct ablation studies to investigate the contribution of each major component in the proposed framework. Specifically, we analyze the effects of real experience integration, multi-objective world model learning, temporal-difference target design, and stochastic latent representation. The complete results are summarized in Table~\ref{tab:ablation}, where return is the average return of 50 trajectories.

\begin{table}[htbp]
\caption{Ablation study results across experiment configurations.}\label{tab:ablation}
\centering
\begin{tabular}{cccccc}
\toprule
\textbf{ID} & $H$ & \textbf{Real Data} & \textbf{Gradient Heads} & \textbf{TD Target} & \textbf{Return} \\
\midrule
A1 & 1  & \ding{55} & decoder $+$ reward $+$ cont & 1-step       & 167.3 \\
A2 & 1  & \ding{51} & decoder $+$ reward $+$ cont & 1-step       & 226.2 \\
A3 & 5  & \ding{51} & decoder $+$ reward $+$ cont & $n$-step & \textbf{371.4} \\
A4 & 5  & \ding{55} & decoder $+$ reward $+$ cont & $n$-step & 347.5 \\
A5 & 5  & \ding{51} & reward $+$ cont             & $n$-step & $-$5.6 \\
A6 & 5  & \ding{51} & decoder only                & $n$-step & 280.8 \\
A7 & 5  & \ding{51} & decoder $+$ reward $+$ cont & 1-step       & 331.6 \\
A8 & 5  & \ding{51} & decoder $+$ reward $+$ cont & $n$-step & 303.4 \\
\bottomrule
\multicolumn{6}{c}{\footnotesize \textit{Note.} A8 uses a continuous Gaussian stochastic latent state; all others use discrete latents. \ding{51} = enabled, \ding{55} = disabled.} \\
\end{tabular}
\end{table}

\textbf{Effect of real and predicted data combination.}
We first evaluate whether real experiences are necessary when training SAC with predicted trajectories. As shown in Table~\ref{tab:ablation}, removing real samples decreases performance for both short and long horizon rollout settings. With $H=1$, training solely on predicted data (A1) achieves limited performance, while incorporating real transitions (A2) leads to clear improvement. A similar trend is observed for $H=5$ (A3 vs.\ A4), although the performance gap becomes smaller because longer rollouts generate more synthetic samples. These results demonstrate that real experiences provide essential grounding signals, while predicted trajectories effectively improve data efficiency by augmenting the training distribution.

\textbf{Effect of multi-objective world model learning.}
We further investigate the influence of different RSSM supervision signals by selectively detaching the gradient flow from individual prediction heads. Detaching the observation reconstruction gradient causes a severe performance degradation (A5), indicating that reconstruction provides an essential regularization signal for learning meaningful latent representations. Although the reward and continuation prediction tasks directly relate to control objectives, their gradients alone are insufficient to maintain a useful representation space. In contrast, detaching reward and continuation gradients while retaining only the reconstruction signal results in moderate performance degradation (A6), suggesting that task-oriented prediction gradients provide additional guidance by encouraging the latent dynamics to preserve decision-relevant information.

\textbf{Effect of temporal target estimation.}
We compare the proposed $n$-step targets for predicted trajectories (A3) with conventional one-step TD targets (A7). The results show that n-step targets achieve better performance, demonstrating that latent rollouts contain useful future information beyond immediate transitions. By propagating rewards over multiple predicted steps, the critic can obtain more informative value estimates and better exploit the learned dynamics.

\textbf{Effect of stochastic latent representation.}
Finally, we compare the default discrete stochastic latent representation (A3) with a continuous Gaussian latent variable (A8). Replacing the discrete latent state with a continuous Gaussian distribution reduces the final performance, suggesting that the discrete stochastic representation provides a more expressive and stable latent space for modeling complex driving dynamics. This observation is consistent with previous findings that categorical latent states can better capture multi-modal future transitions in partially observable environments.

\subsection{Analysis of Model Bias in Rollouts}

To investigate how model bias accumulates during rollouts, we perform a trajectory-level comparison between real and predicted transitions. Specifically, we randomly sample 512 trajectories from the replay buffer and initialize rollouts from the same real states. The world model then generates latent rollouts using exactly the same action sequences as those executed in the real environment. This enables a step-wise comparison between predicted and real trajectories without introducing policy distribution mismatch.

\begin{figure}[htb]
\centering
\includegraphics[width=0.9\linewidth]{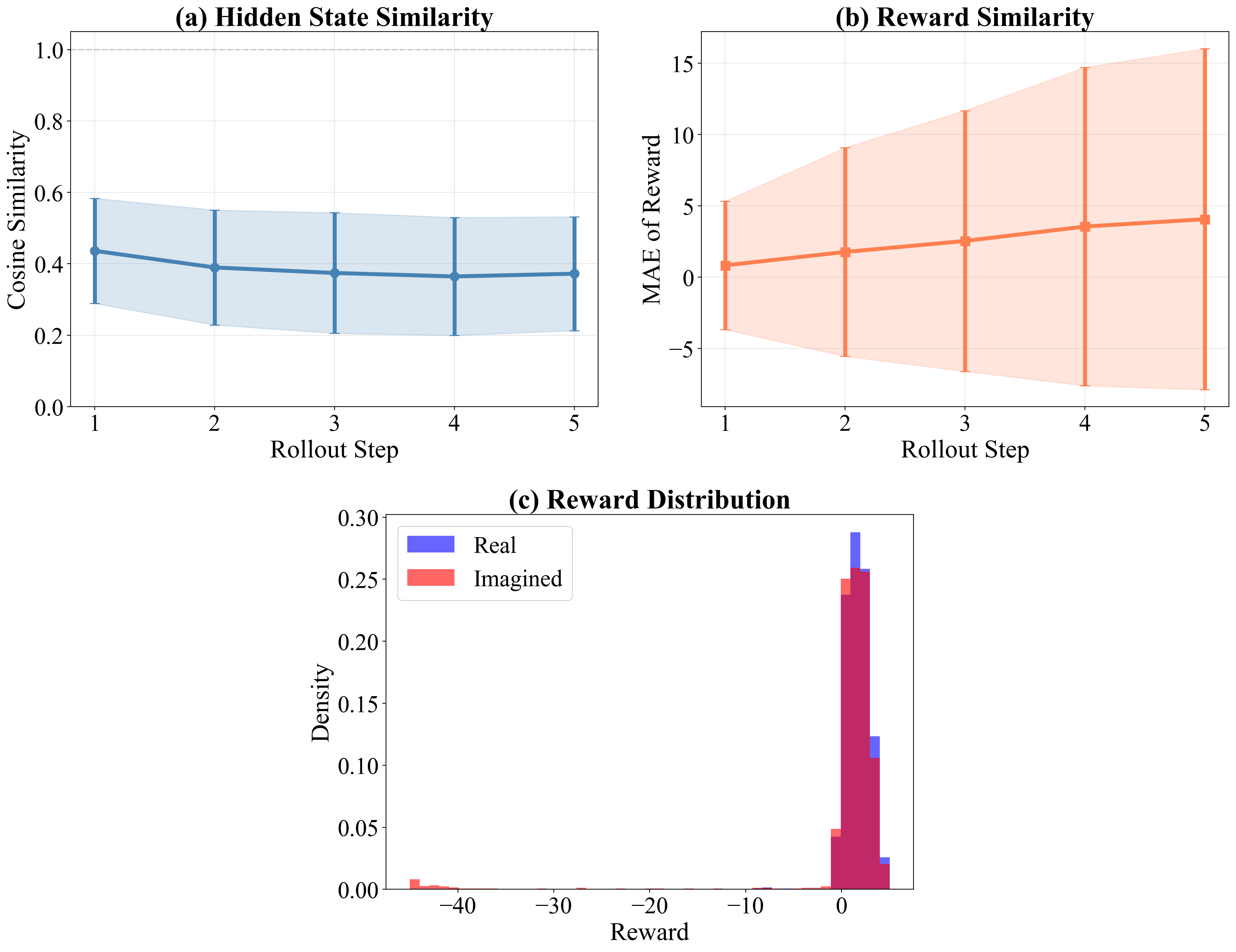}
\caption{Comparison between real and predicted trajectories: latent representation similarity, reward prediction error, and reward distribution over rollout depth.}\label{fig:real_vs_imag}
\end{figure}

Figure~\ref{fig:real_vs_imag} presents the discrepancy between real and predicted trajectories in terms of latent representation similarity, reward prediction error, and reward distribution. The cosine similarity between real and predicted hidden states gradually decreases from 0.438 at the first predicted step to 0.371 after five steps. This indicates that the latent dynamics remain consistent with the real environment in short horizons but gradually accumulate prediction errors during longer rollouts.

Similarly, the reward prediction error increases with rollout depth. The mean absolute error grows from 0.855 at the first step to 3.698 at the fifth step, representing more than a four-fold increase.  Meanwhile, the variance of the predicted rewards is increasing, indicating that uncertainty grows as the rollout process extends.

Further analysis of reward distributions reveals that the world model accurately captures frequent driving rewards around zero, which correspond to normal lane-following and vehicle-control behaviors. However, discrepancies emerge in sparse negative reward regions. The predicted distribution assigns higher probability to extreme negative rewards, indicating that the world model tends to overestimate the occurrence of safety-critical failures such as collisions or off-road events. This conservative prediction behavior results from the difficulty of accurately modeling rare transition events with limited real interactions.

These results demonstrate that latent rollouts provide reliable short-term experience augmentation, while prediction errors gradually accumulate with longer rollouts. This explains the observed trade-off between rollout horizon and policy performance: moderate horizons improve data efficiency by providing additional training signals, whereas excessive rollouts introduce biased value estimation and limit policy optimization.

\section{Conclusion}\label{sec:conclusion}

This paper proposed Dreamer-SAC, a model-based off-policy reinforcement learning framework for autonomous driving that combines the data efficiency of latent world models with the stability and experience reuse capability of off-policy learning. Unlike conventional Dreamer-style approaches that optimize policies solely using model-generated rollouts, the proposed framework jointly leverages real driving experiences and latent rollouts through SAC optimization. By generating short-horizon rollouts from real trajectory states and incorporating n-step target estimation, Dreamer-SAC improves policy learning while reducing the adverse effects of accumulated model errors. Experiments conducted in the MetaDrive autonomous driving environment demonstrate that the proposed approach consistently outperforms representative reinforcement learning baselines, including DreamerV3, SAC, and PPO. Further ablation studies verify the importance of real experience integration, multi-objective RSSM learning, and n-step target estimation. The results also reveal an inverted-U relationship between rollout horizon and policy performance, indicating that short-horizon rollouts provide an effective balance between additional training signals and model bias.

Future work will focus on developing adaptive rollout strategies that dynamically adjust rollout length according to model reliability and uncertainty. In addition, evaluating the proposed framework with larger-scale real-world driving datasets and more diverse traffic scenarios will be essential for further improving the practicality and generalization capability of model-based reinforcement learning for autonomous vehicles.

\bibliographystyle{unsrtnat}
\bibliography{trb_template}

\end{document}